\documentclass[journal]{IEEEtran}
\usepackage{cite}

\ifCLASSINFOpdf
\else
\fi
\usepackage{graphics} 
\usepackage{epsfig} 
\usepackage{newtxtext,newtxmath} 
\usepackage{times} 
\usepackage[font=scriptsize]{caption}
\usepackage{subcaption}
\usepackage{amsmath} 
\usepackage[utf8]{inputenc}
\usepackage[T1]{fontenc}
\usepackage{mathtools}
\usepackage{enumitem}
\usepackage{xcolor}
\usepackage{float}
\usepackage{tabularx}
\newcolumntype{Y}{>{\centering\arraybackslash}X}

 \usepackage{hyperref}
\usepackage{xcolor}
\usepackage{comment}

\usepackage{xcolor,colortbl}

\begin{document}
%
\title{Design, Modeling, and Redundancy Resolution of Soft Robot for Effective  Harvesting}
%
%
%

\author{Milad Azizkhani, Anthony L. Gunderman, Alex S. Qiu, Ai-Ping Hu, Xin Zhang, and Yue Chen
\thanks{This research is funded by Georgia Tech IRIM seed grant,  form-the-team grant, and faculty startup grant. Corresponding author: Yue Chen. }
\thanks{Milad Azizkhani and Anthony L. Gunderman are with the Georgia Tech Institute for Robotics and Intelligent Machines (IRIM), Atlanta, GA, 30313, USA {\tt\small [mazizkhani3, agunderman3]@gatech.edu}}
\thanks{Alex S. Qiu is with the school of mechanical engineering, Georgia Institute of Technology, Atlanta, 30338, USA, {\tt\small aqiu34@gatech.edu}}
\thanks{Ai-Ping Hu is with the intelligent sustainable technologies division, Georgia Tech Research Institute, 640 Strong, Atlanta, 30318 USA, {\tt\small ai-ping.hu@gtri.gatech.edu}}
\thanks{Xin Zhang is with the Department of Agricultural and Biological Engineering, Mississippi State University, Starkville, MS, 39762, USA, {\tt\small xzhang@abe.msstate.edu}} 
\thanks{Yue Chen is with IRIM, and with the Biomedical Engineering Department, Georgia Institute of Technology/Emory, Atlanta, 30313, USA, {\tt\small yue.chen@bme.gatech.edu}}
\thanks{This work has been submitted to the IEEE for possible publication. Copyright may be transferred without notice, after which this version may no longer be accessible.}
}

%
%

\markboth{}%
{Shell \MakeLowercase{\textit{et al.}}: Bare Demo of IEEEtran.cls for IEEE Journals}
%



\maketitle

\begin{abstract}

Blackberry harvesting is a labor-intensive and costly process, consuming up to 50\% of the total annual crop hours. This paper presents a solution for robotic harvesting through the design, manufacturing, integration, and control of a pneumatically actuated, kinematically redundant soft arm with a tendon-driven soft robotic gripper. The hardware design is optimized for durability and modularity for practical use.
The harvesting process is divided into four stages: initial placement, fine positioning, grasp, and move back to home position. For initial placement, we propose a real-time, continuous gain-scheduled redundancy resolution algorithm for simultaneous position and orientation control with joint-limit avoidance. The algorithm relies solely on visual feedback from an eye-to-hand camera and achieved a position and orientation tracking error of 0.64\textpm{0.27} mm and 1.08\textpm{1.5}$^{\circ}$, respectively, in benchtop settings.
Following accurate initial placement of the robotic arm, fine positioning is achieved using a combination of eye-in-hand and eye-to-hand visual feedback, reaching an accuracy of 0.75\textpm{0.36} mm. The system's hardware, feedback framework, and control methods are thoroughly validated through benchtop and field tests, confirming feasibility for practical applications.


\end{abstract}


%
\IEEEpeerreviewmaketitle

\section{Introduction}
Blackberries are a small, delicate fruit that require hand-harvesting to minimize damage \cite{gunderman2022tendon, tomlik2010morphology}. However, there are several compounding factors that negatively influence profitability and marketability. These include a short annual harvesting window (31 days), shortage of available labor, which is strongly influenced by wage and local labor supply, and labor skill \cite{hall2017blackberries}. Moreover, the labor used during the annual harvesting window accounts for 50\% of the total hours spent on the crop annually \cite{hall2017blackberries}. To mitigate labor costs, automated harvesting has been proposed using mechanized shakers for fruit such as blueberries \cite{lobos2014effect}. However, blackberries are easily susceptible to damage that negatively impacts marketability, precluding the use of mechanized shakers in fresh-market applications \cite{myers2022determining}. 
The use of rigid robotic systems with rigid grippers have also been proposed \cite{xiong2018design,avigad2020robotic,shah2019design,de2018development}. However, these systems are often limited by: (i) druplet and stem damage induced by the rigid manipulators and grippers, (ii) strictly greenhouse implementation, and (iii) cost-prohibitive robotic systems.

An alternative to rigid robotic systems are soft robotic systems. Nature-inspired soft robots are characterized by the use of elastic materials in their design and their continuum nature for biological emulation \cite{ren2021biology}. Different from their traditional rigid robotic counterparts, these systems are capable of enabling dexterity, safety, and flexibility in their motions that ensure safe interaction with the environment while augmenting task versatility \cite{laschi2016soft,rus2015design}. Soft robotic arms with soft grippers provide a potential novel solution for low-cost blackberry harvesting by (i) reducing the cost of robotic implementation compared to the conventional manual approach, (ii) mitigating berry and stem damage caused by their rigid robotic counterparts, and (iii) enhancing maneuverability in confined spaces with their infinite degrees of freedom. However, despite their many benefits, soft robot position and orientation control remains an elusive endeavor due to the continuum, nonlinear, and uncertain characteristics of soft robotic arms.

The control problem of soft robots has been addressed previously via model-based and data-driven approaches. In model-based approaches, Cosserat rod theory \cite{renda2014dynamic} and finite element methods \cite{largilliere2015real} provide highly accurate evaluations of the kinematics and dynamics of soft robotics. However, these models are computationally expensive and pose significant challenges for real-time implementation. Model-free (i.e. data-driven) approaches have been proposed as an alternative to inverse kinematic learning of continuum manipulators. Giorelli \emph{et.al} proposed a feed-forward neural network algorithm that solves the inverse kinematics/statics of a 2D soft robotic arm by identifying a mapping from task space to actuator space as a substitute for the model-based Jacobian inverse algorithm to compensate for inaccuracy and time inefficiencies \cite{giorelli2015neural}. Thuruthel \emph{et al.} calculated the inverse kinematics solution of a soft robotic arm through training a feed-forward neural network which mimics
the resolved-rate algorithm for generating control outputs (i.e. incorporating desired position, current position, and current actuator variables) \cite{george2017learning}. Satheeshbabu \emph{et al.} implemented a 
reinforcement learning algorithm (DDPG) for evaluating the inverse kinematics of the robot \cite{satheeshbabu2020continuous}. However, model-free approaches require extensive offline data collection and learning, and cannot guarantee generalization and system stability.  

To resolve the aforementioned issues, research groups have implemented simplifying assumptions, such as piece-wise constant curvature, to enable real-time control \cite{godage2015modal,azizkhani2022dynamic2,godage2016dynamics}. 
However, hysteretic effects, pressure dynamics, uncertain and variable stiffness, along with manufacturing uncertainties, such as deviations from piece-wise constant curvature assumption, make the control problem difficult. 
General inverse kinematic controllers embedded with mechanical models \cite{bieze2018finite}, as well as learning approaches \cite{giorelli2015neural,george2017learning,satheeshbabu2020continuous} have been proposed to compensate for these uncertainties. However, they either require rigorous data-collection and learning or computationally expensive models, which make their implementation challenging in real world applications.
Efficient harvesting requires simultaneous closed-loop control in both orientation and position to ensure proper alignment of the gripper with respect to the berry during the harvesting procedure. To the best of our knowledge, closed-loop simultaneous position and orientation control of redundant soft pneumatic robot arms has not been previously addressed in literature. The closest related work can be seen in \cite{godage2015modal}, where the algorithm was performed using a constrained optimization algorithm in MATLAB using the "fmincon" algorithm. However, this study is limited to simulation cases, and the kinematic model was assumed to be exact without any uncertainties. Moreover, the inverse kinematic solution was found in an open-loop manner, relying entirely on the model.

The proposed soft robotic harvesting system and continuous, closed-loop control algorithm aim to enable field implementation of blackberry harvesting. This paper's contribution is summarized as follows:
\begin{itemize}
    \item The design and implementation of a soft robotic system for blackberry harvesting.
    \item A kinematic control algorithm for simultaneous closed-loop position and orientation control for a kinematically redundant, soft, pneumatically actuated arm using only visual feedback from an eye-to-hand camera.
    \item Achieving stable closed-loop control performance using a gain-scheduled algorithm in the redundancy resolution algorithm that enables simultaneous position and orientation control while avoiding joint limits.  
    \item Fine positioning control achieved using primary feedback from an eye-in-hand camera and auxiliary information from an eye-to-hand camera.
    \item Satisfactory performance achieved in the presence of kinematic uncertainties, hysteresis, and nonlinearities of the system at slow speeds.
\end{itemize}

The remainder of the paper is organized as follows. Section \ref{sec: system overview} presents an overview of the system hardware, including the soft robotic arm, soft robotic gripper, actuation system, and the tracking system. Section \ref{sec: control and task synchronization} discusses the kinematic modeling and control of the system. Section \ref{sec: Experimental Results} presents the experimental methods and results. The paper is concluded in Section \ref{sec: Conclusion}.

\section{Soft Robot Hardware Development}
\label{sec: system overview}
\begin{figure}[t]
    \centering
    \includegraphics[width = 1.00\linewidth]{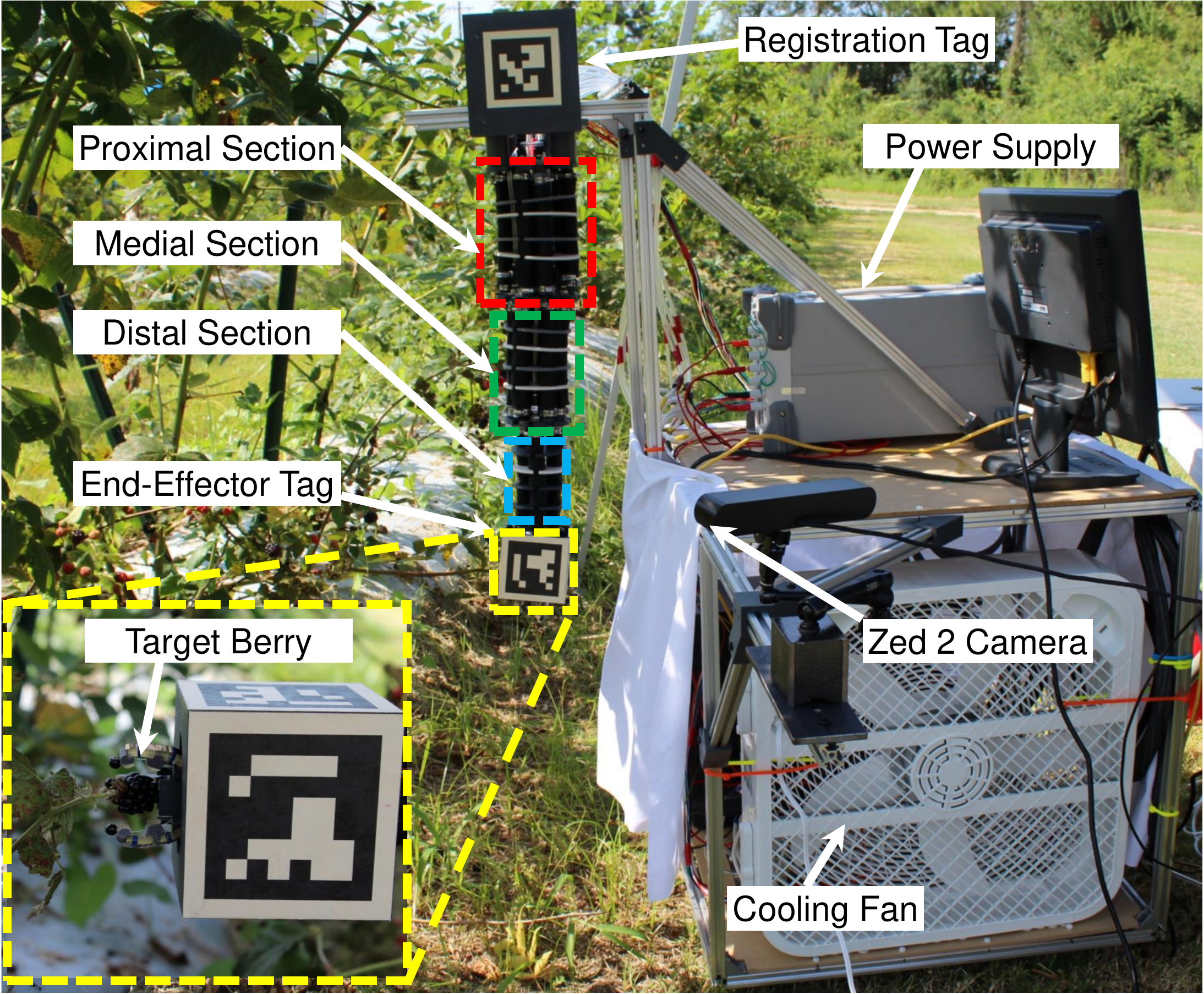}
    \caption{Proposed soft robotic harvesting platform, which consists of the soft robotic arm, soft gripper, tracking, control and actuation system. Red, green, and blue depict proximal, medial, and distal sections, respectively.
    }
    \label{fig:robotc_system}
    \vspace{-2mm}
\end{figure}

The robotic system (Fig. \ref{fig:robotc_system}) can be divided into four different subsystems: 
(i) the soft robotic arm, (ii) the soft robotic gripper, (iii)  the tracking system, and (iv) the robot control mechatronics. These subsystems are placed on a truss-based  chassis made from 20-mm t-slot (5537T101, McMaster-Carr, USA). 
The chassis is constructed in a manner that enables easy mounting to a potentional mobile platform using sliding brackets (HK1000-V2, SuperDroid Robots, USA). In the following sections, we will present each subsystem in detail. 

\subsection{Soft Robotic Arm}
\subsubsection{Soft Actuator Design}
The actuator design is an extension-based pneumatic artificial muscle (Fig. \ref{fig:arm_schematic} A-B). The core of the muscle consists of a silicone segment cut to a length of 15-cm (4G-60518-latex, Feelers, USA). The outer mesh of the muscle consists of a 5/8" expandable braided sleeve cut to 45-cm (B00ZCNTIRQ, Electriduct, USA). The muscle is plugged using a stereolithography (SLA) printed (Form 3B, FormLabs, USA) top plug and base adapter.  The base adapter and top plug have threads cut onto the end using a 3/8"-24 die. These threads are used for attaching a clamping nut to the pneumatic artificial muscle. The base adapter is designed to permit the use of a push-to-connect pneumatic fitting (B09DCRF4F9, Hotop, USA). The silicone core and outer mesh are connected to the base adapter and top plug using pneumatic clamps (B078BRJK8Z, LOKMAN, China). A piece of sacrificial tubing (Z20210816W25, VictorsHome, USA) is placed between the outer mesh and the clamp to prevent premature silicone failure from the sharp edges of the pneumatic clamp.

\subsubsection{Multi-Section Soft Arm Design}
\begin{figure}[t]
    \centering
    \includegraphics[width = 1\linewidth]{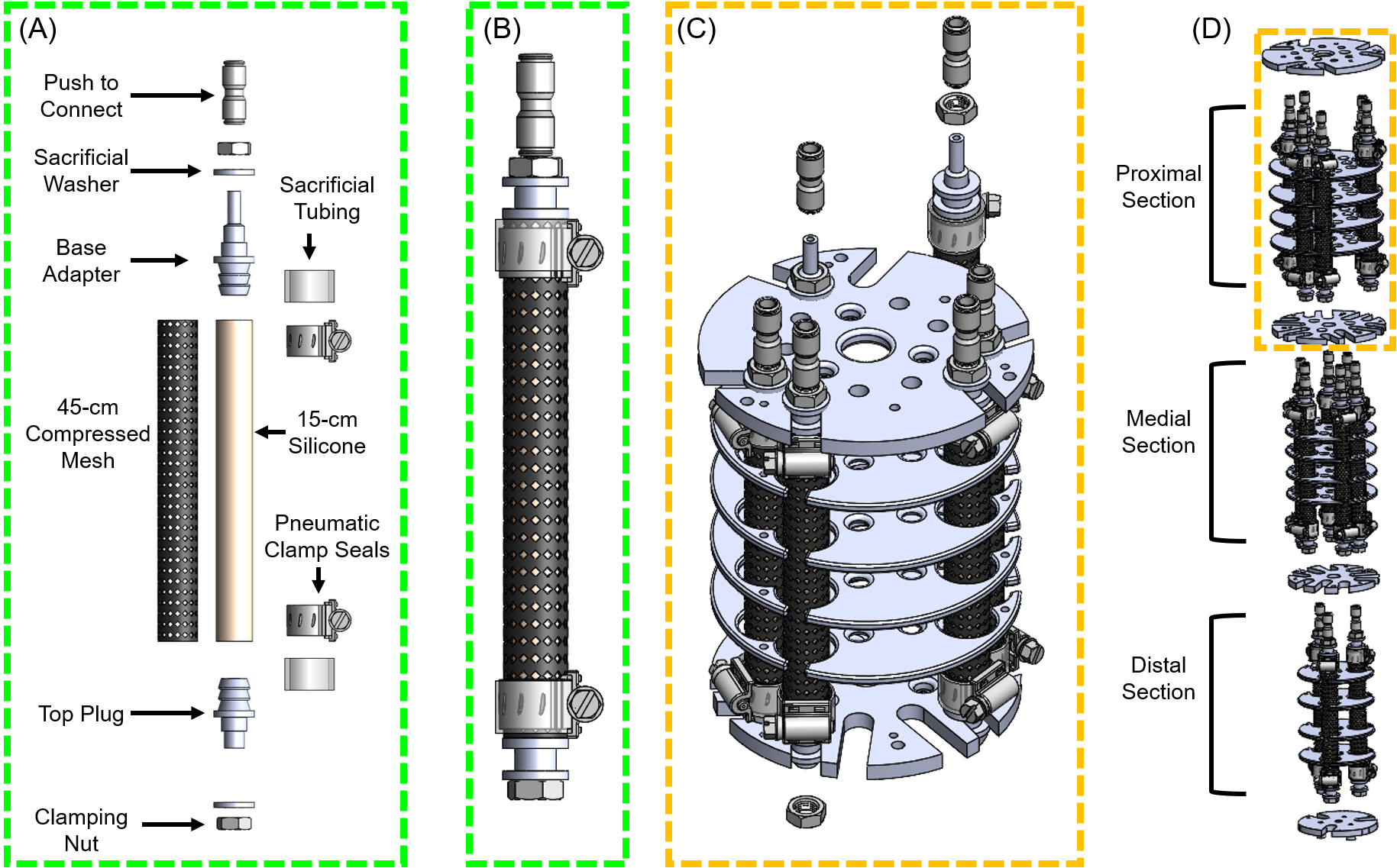}
    \caption{
(A) Single actuator design in its expanded and (B) assembled configuration. (C) The exploded view of the assembly procedure in the proximal section. (D) The exploded view of the soft robotic arm shows its composition of three sections, namely the proximal, medial, and distal sections. The proximal and medial sections consist of three pairs of two actuators, while the distal section is comprised of three actuators.} 
    \label{fig:arm_schematic}
    \vspace{-2mm}
\end{figure}

The soft robotic arm design consists of three modular sections, which we refer to as the proximal, medial, and distal sections, corresponding to increasing distance from the base (Fig. \ref{fig:robotc_system} and \ref{fig:arm_schematic}).
As the segments approach the base (from the end-effector), the actuator spacing from the center-line increases to enhance the stiffness. The proximal and medial sections incorporate three pairs of two actuators each, which are connected to regulators, resulting in a strength-augmented actuator system. Conversely, the distal section features only three individual actuators, which impart greater dexterity in this segment.
The design of each section of the soft robotic arm features a 60° offset from its neighboring sections to facilitate the routing of the pneumatic transmission lines. This design also includes slots to enable quick exchange of actuators in the event of failure. The robot arm has intermediate plates that run along the actuators to preserve the constant curvature assumption. These plates have been designed with gaps to simplify the removal of the actuators.

\subsection{Soft Gripper}
The soft gripper is developed based on our prior work  \cite{gunderman2022tendon} with the primary focus to enhance the system compactness and incorporate the berry ripeness detection module. 
The gripper consists of three tendon-driven soft  fingers.
The gripper fingers are made of flexible urethane (VytaFlex 60, Smooth-On, USA) and possess a steel sheet to serve as the backbone (9293K111, McMaster-Carr, USA). The fingers are actuated using a servo motor (SG90, TowerPro, USA) that uses rotary motion to retract the finger tendons (BGQS8C-15, Berkley, USA). This compact, light-weight design was motivated by a desire to reduce the weight compared to our prior work \cite{gunderman2022tendon}. In the palm of the gripper, a near-infrared (NIR) based blackberry ripeness sensor is incorporated that detects berry ripeness by utilizing the berry reflectance characteristic. The detailed description of the ripeness detection unit design can be found in \cite{qiu2023gripper}.

\subsection{Tracking System}
\label{subsec: Tracking system}
The tracking system utilizes two different feedback modalities: 1) eye-to-hand camera (C1) and 2) eye-in-hand camera (C2). The eye-to-hand camera (ZED2, STEREOLABS, USA) is used for online robot registration and end-effector localization.
The eye-to-hand camera is placed approximately 1.2 m from the robot base and is oriented in a way to ensure the robot workspace is within the field of view (FOV). The eye-in-hand camera (DPNKJ0015, DEPSTECH, USA) is used when the berry is occluded by the end-effector in the eye-to-hand camera FOV. 
The eye-in-hand camera is placed in the center of the end-effector and the cable is routed through the middle of the soft robotic arm. 

Robot registration and end-effector tracking are performed using ArUco markers. One ArUco tag was attached to the base for online coordinate frame registration of the robot base. End-effector tracking in both position and orientation is obtained via a cube covered by four distinct ArUco tags, as shown in Fig. \ref{fig:robotc_system}. Based on the  ArUco tags feedback, the robot end-effector position can be obtained via basic homogeneous transformation. 
We would like to note that the 80mm ArUco markers are used in the preliminary design  for effective tracking. Additionally, a low-pass filter with a cut-off frequency of $10 \ rad/sec$ is used to suppress the potential signal noise.

\subsection{Robot Control Mechatronics}
The soft robotic arm and gripper control system is developed on the Simulink xPC target machine. The soft robotic arm is controlled using nine proportional pressure regulators (ITV1031-21N2BL4, SMC Corporation). The xPC target machine is connected to an analog output board (PCI-6703, National Instruments) which provides  16-bit control voltages (0-5 V) to the pressure regulators. The pressure regulators are connected to the soft robotic arm via nine 3-m pneumatic air hoses (5233K52, McMaster-Carr, USA). The Simulink xPC targeting machine and proportional pressure regulators are mounted to the bottom of the robotic chassis and  cooled using a portable box fan as depicted in Fig. \ref{fig:robotc_system}. The power supply provides  24 volts for the pneumatic valves, and 5 volts for the servo motor and Arduino.

\section{Robot Controller Development}
\label{sec: control and task synchronization}

The control strategy for berry harvesting involves two main tasks: (i) positioning the end-effector with proper distance and orientation so that the berry is visible to the eye-in-hand camera, and (ii) approaching the target until the berry is within the gripper's grasp workspace. Both tasks require generalized position and orientation control, but the second task also involves additional visual feedback from the eye-in-hand camera to ensure accurate positioning for grasping, particularly when the end-effector occludes the target.

This section begins with a review of the kinematic model of the soft robot and proceeds to develop a kinematic control strategy based on the standard resolved-rate algorithm. The discussion then covers gain-scheduled modification and joint-limit avoidance, which are critical for ensuring stable robot control. Additionally, a fine positioning process is described that uses the eye-in-hand camera as primary feedback to enable precise control as the robot approaches the target. Finally, the overall workflow of the harvesting process is outlined with an emphasis on achieving stable transitions between each step.

\subsection{Robot Kinematic Modeling and Control}
\begin{figure}[t]
    \centering
    \includegraphics[width = 1.00\linewidth]{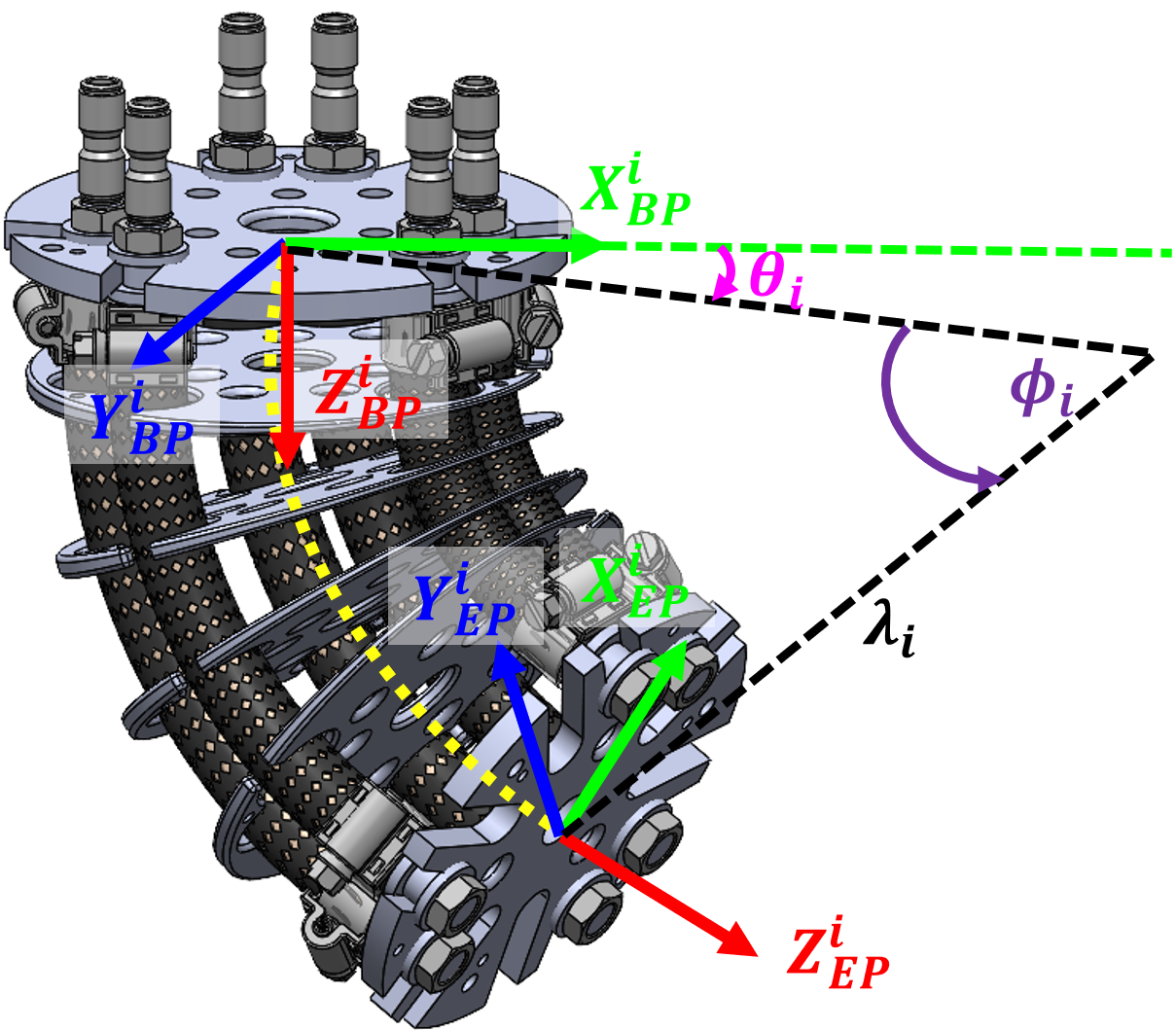}
    \caption{The general notation for the forward kinematics of section $i$, which describes the transformation of the end plate (EP) with respect to the base plate (BP).
    \label{fig: ForwardKinematicsSchematics}
    }
\end{figure}

To mathematically describe the kinematics of soft robots, two mappings are necessary: one that maps the actuator variables to the robot's configuration space, and another that maps the configuration space to the task space. For soft robots, the actuator variables are the changes in length, $l_{i,j}$, of each actuator, where $i \in {1, 2, 3}$ represents the section number and $j \in {1, 2, 3}$ represents the actuator number within that section. The initial length of all actuators is denoted as $L_0$.

In task space, the $X_i$ axis is oriented towards the first actuator, the $Z_i$ axis is perpendicular to the base plate and extends towards the end plate of the section, and the $Y_i$ axis is defined using the right-hand rule. The actuators (or pairs of actuators in proximal and medial sections) are positioned $120^\circ$ apart from each other, with a radius of $r_i$ from the center of the base. Fig. \ref{fig: ForwardKinematicsSchematics} shows the forward kinematics schematics diagram for section $i$.

Leveraging the piece-wise constant curvature assumption  \cite{azizkhani2022dynamic}, the mapping from actuator-space to configuration-space of section $i$ 
is formulated as
\begin{equation}
    \begin{aligned}
            s_{i} &= \sqrt{{l_{i,1}}^2+{l_{i,2}}^2 + {l_{i,3}}^2 -l_{i,1}l_{i,2} - l_{i,2}l_{i,3} - l_{i,1}l_{i,3}}, \\   \phi_{i} &= \frac{2s_{i}}{3r_{i}},\
            \lambda_{i} = \frac{(3L_{0}+l_{i,1}+l_{i,2}+l_{i,3})r_{i}}{2s_{i}}, \\ \theta_{i} &= \arctan\left({\frac{\sqrt{3}(l_{i,3}-l_{i,2})}{l_{i,2} + l_{i,3} - 2l_{i,1}}}\right)
    \end{aligned}
    \label{eq: FKin}
\end{equation}
where $\lambda_{i} \in (0,\infty)$, $\phi_{i} \in [0, 2\pi)$, and $\theta_{i} \in [-\pi,\pi]$ represent the radius of curvature, bending angle of the arc, and bending angle with respect to defined $X_{i}$ axis, respectively. Next, the mapping from configuration-space to task-space of section $i$ is derived as follows
\begin{equation}
\begin{aligned}
    T_i(\boldsymbol{q}_i) =& \text{Rot}_{z}(\theta_i)\text{Trans}_{x}(\lambda_i)\text{Rot}_{y}(\phi_i)\\ &\text{Trans}_{x}(-\lambda_i)\text{Rot}_{z}(-\theta_i)
\end{aligned}
\end{equation}
where $\text{Rot}_k$, and $\text{Trans}_k$ are rotational and translation homogeneous transformations defined as in \cite{spong2008robot}, with respect to the $k$ axis. Based on the geometry of the soft robotic arm, the forward kinematics of end-effector (EE) frame ($T_{EE}$) with respect to the base frame is derived as follows
\begin{equation}
\begin{aligned}
    T_{EE} =& \prod_{i=1}^{n=3}[T_{i}(\boldsymbol{q}_i)\text{Trans}_{z}(b_{i})\text{Rot}_{z}(\beta_{i})]
\end{aligned}
\end{equation}
where $\text{Trans}_{z}(b_{i})$ and $\text{Rot}_{z}(\beta_{i})$ are the linear and angular offset of frame $(i+1)$ with respect to frame $(i)$, respectively. 
The last offsets ($b_3$, and $\beta_3$) are the gripper offset with respect to end plate of the distal section. 

After kinematic evaluation, we calculate the robot Jacobian based on the kinematics model developed above.  The $m^{th}$ column of linear velocity Jacobian is derived as
\begin{equation}
    J_v^m = \frac{\partial p_{EE}}{\partial q_m}
\end{equation}
where $m$ is the $m^{th}$ member in actuator variables vector $\boldsymbol{q} = [l_{1,1},l_{1,2},l_{1,3},l_{2,1},l_{2,2},l_{2,3},l_{3,1}l_{3,2},l_{3,3}]^T$, and $p_{EE}$ is the end-effector position in base frame. The  $m^{th}$  column of angular velocity is also derived as
\begin{equation}
    J_\omega^m = (\frac{\partial R_{EE}}{\partial q_m} R_{EE}^T)^{\vee}
\end{equation}
where $R_{EE}$ is the end-effector rotation matrix with respect to the base frame, and $()^\vee$ is defined based on \cite{murray2017mathematical}. Thus, the complete Jacobian can be represented as
\begin{equation}
    J = \begin{bmatrix}
    J_v \\
    J_\omega
    \end{bmatrix}
\end{equation}
After calculating the Jacobian, the general resolved-rate algorithm in discrete format can be described as follows
\begin{equation}
\begin{aligned}
    &\dot{\boldsymbol{q}} = \alpha J(\boldsymbol{q}(t))^{+} (P_{d}-P_{c})/{\Delta t}\\
    &\boldsymbol{q}(t+ \Delta t) = \dot{\boldsymbol{q}}{\Delta t} + \boldsymbol{q}(t)
    \label{eq: res_g}    
\end{aligned}
\end{equation}
where $P_{d}$ and $P_{c}$ denote desired and current pose of the end-effector with respect to the base frame, respectively, and $\alpha$ is a tuning constant coefficient which defines the speed change. Note that $\alpha$  is crucial for the convergence of the algorithm. If $\alpha$ is set to be a large value, then the solution will oscillate around the optimal solution. It should be mentioned that the algorithm will iterate until it reaches  the predefined error thresholds.

\subsection{Robot Redundancy Resolution}
\label{sec: Redundancy Resolution}
The direct implementation of \eqref{eq: res_g} in its general form often leads to instability due to the difference in units between position and orientation \cite{siciliano1990closed}. A more practical approach is to use two separate sets of gains $\alpha$ for position and orientation control in \eqref{eq: res_g}. However, even in the modified approach, it still leads to problematic behavior in the proposed soft robotic arm control experiments. This is potentially due to the geometrically coupled structure of the soft arm, which can be considered as a hybrid serial-parallel robot. In classical serial manipulators, such as PUMA 560 or UR robot, the simultaneous control of position and orientation is easily achieved via the wrist-based end effector that permit position and orientation decoupling \cite{paul1983kinematics}. However, this wrist joint is lacking in the proposed soft robotic arm, leading to the challenges in robot control.

In this section, we will detail the computational framework to obtain a feasible solution for soft robot control. Note that the physical limitation of each actuator also needs to be taken into account during calculation. The pneumatic artificial muscles can only be elongated (used in this study) or contracted, depending on the outer mesh braid angle \cite{liu2003fiber}. Thus, joint-limits of the actuators need to be considered (i.e. $0<l_{i,j}<l_{i,j}^{max}$). 
Thus,  we propose the redundancy resolution algorithm \cite{nakamura1987task} to achieve the simultaneous position and orientation control (geometrical constraint) and satisfy the joint limit constraints (physical constraint) in soft continuum arms. The control goal is divided into three sub-tasks: position control (Task 1), orientation control (Task 2), and joint-limit avoidance (Task 3).

\subsubsection{Position and Orientation Control}

Following the differential kinematics, the robot position error $e_p$  and orientation error $e_\zeta$  with respect to joint velocities can be written as

\begin{equation}
\begin{aligned}
\text{Task 1:}  \quad e_p &= J_v \dot{\boldsymbol{q}}\\
\text{Task 2:} \quad e_\zeta &= J_\omega \dot{\boldsymbol{q}}
\label{eq: task pri}
\end{aligned}
\end{equation}

In accordance with the task prioritization redundancy resolution formulation outlined in \cite{nakamura1987task}, the solution for fulfilling the first task while incorporating the solution for the second task is given as follows
\begin{equation}
\dot{\boldsymbol{q}} = J_v^{+}e_p + (I-J_v^{+}J_v)\eta_1
\label{eq: task1}
\end{equation}
where $\eta_1$ is calculated to accommodate the secondary task by taking into account the null space projection of the first task, $(I-J_v^{+}J_v)$. Substituting $\dot{\boldsymbol{q}}$ in \eqref{eq: task1} into task 2 \eqref{eq: task pri} yields
\begin{equation}
e_\zeta = J_w(J_v^{+}e_p + (I-J_v^{+}J_{v})\eta_1)
\end{equation}
Solving for $\eta_1$ and incorporating it into \eqref{eq: task pri} gives 
\begin{equation}
\dot{\boldsymbol{q}} = J_v^{+}e_p + (I-J_v^{+}J_v)J_\omega^{+}(e_\zeta-J_\omega J_v^{+}e_p)
\label{eq: General RR}
\end{equation}

To ensure consistent and stable behavior, tuning parameters have been incorporated into \eqref{eq: General RR} as follows
\begin{equation}
\dot{\boldsymbol{q}} = \alpha_1 \left( J_v^{+}e_p + \gamma_1(I-J_v^{+}J_v)J_\omega^{+}(e_\zeta-J_\omega J_v^{+}e_p) \right)
\label{eq: red2}
\end{equation}
where $\alpha_1$ and $\gamma_1$ are tuning constants used to stabilize the numerical calculation steps.
Finally, the desired actuator space input for each actuator can be computed using \eqref{eq: red2} for simultaneous position and orientation control.

\subsubsection{Actuator Joint-Limit Avoidance}

As discussed, the joint-limit constraint must be considered in the formulation to generate feasible solutions. To address this issue, we can incorporate redundancy resolution to optimize a cost function, prioritizing tasks 1 or 2 over task 3. The solution for the joint-limit constraint can be expressed as
\begin{equation}
\begin{aligned}
\text{Task 1\&3:}  \quad\dot{\boldsymbol{q}} &=\alpha_2 \left( J_v^{+}e_p + \gamma_2(I-J_v^{+}J_v)\eta_2 \right)\\
\text{Task 2\&3:}  \quad \dot{\boldsymbol{q}} &=\alpha_3 \left( J_{\omega}^{+}e_\zeta + \gamma_3(I-J_\omega^{+}J_\omega)\eta_2 \right)
\end{aligned}
\label{eq: redjoi}
\end{equation}
where $\eta_2$ is the gradient of a joint-limit cost function ($\nabla H$) defined based on the physical constraints of the actuators, which allows us to follow the standard gradient projection method for redundancy resolution. $\alpha_2$, $\alpha_3$, $\gamma_2$, and $\gamma_3$ are tuning parameters that help stabilize the numerical calculations. The joint-limit cost function, as defined in \cite{chan1995weighted}, is as follows
\begin{equation}
H(\boldsymbol{q}) = \sum_{m=1}^{9} \frac{1}{4} \frac{(q_m^{max}-q_m^{min})^2}{(q_m^{max} - q_m)(q_m - q_m^{min})}
\label{eq: cost}
\end{equation}
By incorporating the joint-limit cost function in the redundancy resolution algorithm, the proposed algorithm ensures that the generated solutions consider the joint limits imposed by the soft actuators  are within the feasible range.

\subsubsection{Simultaneous Position and Orientation Control with Joint-Limit Avoidance}
Now it is time to combine the decoupled position and orientation control, with joint-limit avoidance. 
Based on our application, where the gripper should be placed the desired location first, and then orientate towards the proper angle, the priorities of these tasks are defined as (i) position control, (ii) orientation control, and (iii) joint-limit avoidance. To achieve this, similar to \eqref{eq: task1}, we write the redundancy resolution algorithm as follows
\begin{equation}
  \dot{\boldsymbol{q}} = J_v^{+}e_p + (I-J_v^{+}J_v)\eta_3  
   \label{eq: eta3 null space}
\end{equation}
where $\eta_3$ is defined as
\begin{equation}
    \eta_3 = \Tilde{J_v^{+}}(e_\zeta - J_\omega J_v^{+}e_p) + (I - \Tilde{J_v^{+}}\Tilde{J_v})\eta_2
    \label{eq: eta3}
\end{equation}
and $\Tilde{J_v}$  is defined as 
\begin{equation}
    \Tilde{J_v} = J_\omega (I-J_v^{+}J_v)
\end{equation}

Incorporating \eqref{eq: eta3} in \eqref{eq: eta3 null space}, and using proof from \cite{maciejewski1985obstacle, nakamura1987task}, we reach the following equation that considers all three tasks together
\begin{equation}
\text{All Tasks:} \quad    \dot{\boldsymbol{q}} = J_v^{+}e_p + \Tilde{J_v}^{+}(e_\zeta-J_\omega J_v^{+}e_p)
    + (I-J_v^{+}J_v-\Tilde{J_v}^{+}\Tilde{J_v})\eta_2
    \label{eq: complete redundancy resolution without gains}
\end{equation}

Note that equation \eqref{eq: complete redundancy resolution without gains} would not result in stable results due to different magnitudes of each term in the right hand side. We reformulate to system with tuning parameters ($\gamma_4$ and $\gamma_5$)  as follows
\begin{equation}
    \begin{aligned}
    &\dot{\boldsymbol{q}} = J_v^{+}e_p + \gamma_4 \Tilde{J_v}^{+}(e_\zeta-J_\omega J_v^{+}e_p)\\
    &+ \gamma_5 (I-J_v^{+}J_v-\Tilde{J_v}^{+}\Tilde{J_v})\eta_2
    \end{aligned}
    \label{eq: joint control}
\end{equation}

In summary, equation \eqref{eq: joint control} details the proposed soft robotic arm kinematic control, which  relies on the task priority redundancy resolution with the consideration of  joint limit avoidance. 

\subsection{Gain Scheduling Resolved-Rate Algorithm and Parameter Tuning} 
To apply \eqref{eq: joint control} in our system, it is necessary to adjust the parameters to ensure stable and reliable performance. To achieve this, we need to comprehend the function of each term in \eqref{eq: joint control} and their influence on the generated solution. First, we will examine the scenario where resolved-rate control is utilized for position control and joint limit avoidance, prioritizing task 1 over task 3. It is worth mentioning that the same principle applies when task 2 is given priority over task 3. Subsequently, we will expand the discussion to encompass simultaneous position and orientation control with joint limit avoidance.

The joint limit avoidance term in the redundancy resolution, as described in \eqref{eq: redjoi}, influences the solution by performing a null space projection of the primary task. In essence, it modifies the solution vector generated by the conventional resolved-rate control ($J_v^{+}e_p$) and guides the results to a feasible local manifold. It is worth mentioning that the magnitude of the second term $(I-J_v^{+}J_v)\eta_2$ in \eqref{eq: redjoi} is different from that of the first term and can be adjusted through the parameter $\gamma_2$.

The joint limit cost function \eqref{eq: cost} reaches its minimum value when the elongation is at the middle of the actuator bounds, and it diverges to infinity as it approaches the limits. By choosing a negative value for $\gamma_2$, the second term tries to minimize the cost function and push each actuator towards the middle of its physical bounds. As such, it is crucial to balance the joint limit avoidance effects by appropriately tuning the magnitude of $\gamma_2$, to avoid drastically altering the solution while still providing enough modification to push it into a feasible region.
However, it is important to note that the joint limit avoidance effect remains active at all times, even as the robot approaches the desired position or orientation. In a scenario where the robot is close to the desired position with a feasible solution, the magnitude of the $J_v^{+}e_p$ term will be small while the joint limit avoidance magnitude could still be relatively high if the actuator is close to the joint limit. To address this issue, a gain scheduling algorithm is necessary to provide joint limit avoidance when the robot is far from the desired position and cancel out the joint limit avoidance protocol when its effect becomes problematic (i.e., in close proximity to the desired position).
The consideration of joint-limit avoidance in hybrid position and orientation control requires a more sophisticated approach than in the case of only position control. In order to effectively decouple the position and orientation control while ensuring the joint-limits are not exceeded, the algorithm in \eqref{eq: redjoi} is modified as follows
\begin{equation}
\begin{aligned}
    &\dot{\boldsymbol{q}} = \alpha_4 \left[ \gamma_6 J_v^{+}e_p + \gamma_4 \Tilde{J_v}^{+}(e_\zeta-J_w J_v^{+}e_p) \right.\\
    &+ \left. \gamma_5 (I-J_v^{+}J_v-\Tilde{J_v}^{+}\Tilde{J_v})\eta \right]\\   
\end{aligned}
\label{eq: rrra}
\end{equation}
to provide enough flexibility for parameter's tuning and gain scheduling. 

The challenge with tuning the parameters for this hybrid control algorithm lies in balancing the desired position, orientation, and joint limits simultaneously. To achieve this, the algorithm first employs hybrid position and orientation control (Task 1 and 2), with the projection in the null space of the position control Jacobian ensuring decoupled behavior between position and orientation control. To tune the parameters $\gamma_4$ and $\gamma_6$, the robot is instructed to reach a desired pose that can only be achieved in its fully extended region. $\gamma_4$ needs to be set to a lower value compared to $\gamma_6$, and its value needs to be adjusted (similar to position control and joint limit avoidance) to push the solution towards a local manifold where the desired orientation is achievable. The $\gamma_4$ value should not be too large, as its effect should not dominate the primary task. Unlike joint-limit avoidance, the $\gamma_4$ value does not need to be switched since both the position and orientation error are decreasing during the robot's movement.

The next step involves tuning joint-limit avoidance with respect to the hybrid position and orientation control. The null space projection for this task now includes the position and orientation control, but the tuning process is similar to separate position control and joint limit avoidance. We treat the simultaneous position and orientation control as one task and adjust $\gamma_5$ accordingly. The same logic from position control and joint limit avoidance applies, and the switching function should be incorporated to ensure stable and consistent behavior. Once the algorithm has been successfully tuned for stable performance, the next step is to fine-tune the $\alpha_4$ parameter to improve the convergence speed. This involves adjusting the value of $\alpha_4$ such that the robot reaches the desired pose faster while still maintaining stable behavior. The process of tuning $\alpha_4$ can be carried out by incrementally increasing or decreasing its value and observing the effect on the robot's performance. A suitable value of $\alpha_4$ should be one that provides a good balance between convergence speed and stability.

To ensure consistent and robust performance, another factor must also be taken into account. It is imperative that the initial conditions for the actuator variables are within their limits, as this will prevent instability in the algorithm. To consider this, the initial conditions have been carefully chosen to lie within the feasible domain. Additionally, to avoid any singularities in the Jacobian, a small perturbation has been introduced to the initial actuator variables.

\subsection{Visual Servoing  and Overall Workflow}
\label{subsec: vis servo1}

In the initial stage of the berry-harvesting process, the placement of the gripper with respect to the berry can be achieved using eye-to-hand camera feedback. The eye-in-hand camera is also integrated with the gripper design to enable accurate feedback for focal fine control. 
However, solely relying  on eye-in-hand camera feedback will need complex control approaches. Previous studies have addressed the issue of accurate and consistent visual servoing of soft robotic manipulators using adaptive control strategies \cite{ wang2016visual, xu2021visual}, but the resulting formulations are challenging to integrate for general applications.

In this paper, we propose a simple approach that utilizes information from both eye-in-hand and eye-to-hand cameras to transform the error to the base frame, enabling accurate and stable resolved-rate control as described in the previous section. This approach results in a much simpler formulation and requires only minor modifications to achieve accurate results. The use of hybrid measurements from both cameras not only enables accurate initial placement of the gripper, but also helps with precise position control when the end-effector occludes the target. This is particularly useful in the case of blackberry harvesting, where the berries can easily be hidden from the eye-to-hand field of view by the gripper. The combination of measurements from both cameras allows for more robust and accurate control throughout the entire harvesting process.

The transformation of the berry position from eye-in-hand camera to base frame can be achieved as follows
\begin{equation}
    P_{Berry}^{Base} = T_{EE}^{Base}  T_{C2}^{EE} P_{Berry}^{C2}
\end{equation}
where $P_{Berry}^{C2}$ provide the berry position in C2 frame (eye-in-hand camera frame).
During our laboratory experiment, a 12mm ArUco tag is attached to the artificial berry to define the target. 
$T_{C2}^{EE}$ is derived based on point-registration algorithm \cite{fitzpatrick1998predicting} to calculate the homogeneous transformation from $C2$ to $EE$ frame.
$T_{EE}^{Base}$ is defined as follows
\begin{equation}
    T_{EE}^{Base} = T_{C1}^{Base} T_{EE}^{C1}
\end{equation}
where 
$T_{EE}^{C1}$ is found using image-processing algorithm from eye-to-hand camera, and  $T_{C1}^{Base}$ is derived by implementing point-registration.

\begin{figure}
    \centering
    \includegraphics[width = 1\linewidth]{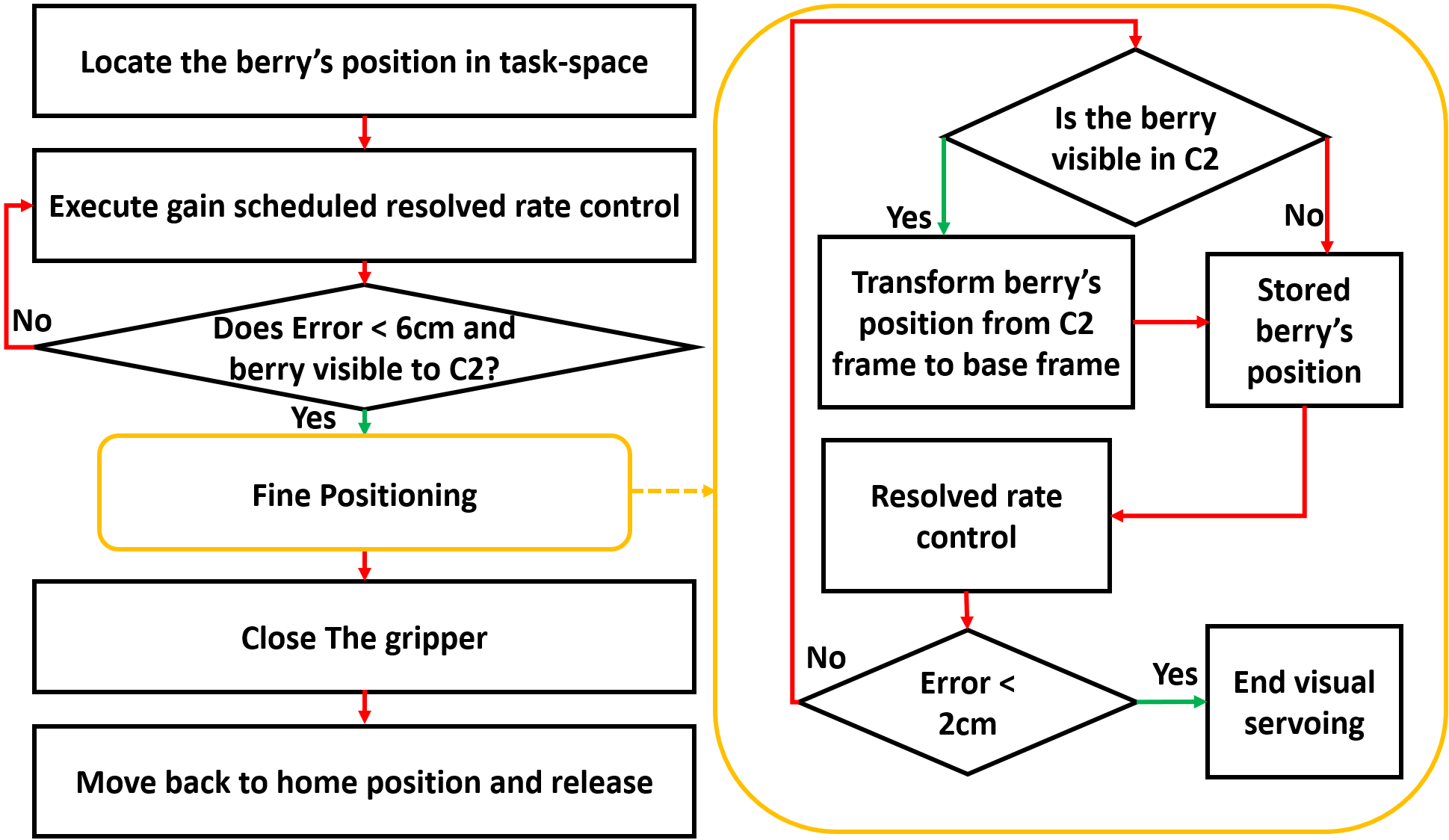}
    \caption{Overall process for effective grasping.}
    \label{fig: vis ser}
\end{figure}

Figure \ref{fig: vis ser} provides a streamlined illustration of this proposed process. The eye-to-hand camera feedback is used to drive the robot towards the target within the predefined distance threshold (6cm). Then the feedback is switched to the eye-in-hand camera, which allow the robot to perform fine gripper control. Note that it is possible that the berry might be occluded by the leafs during the practical scenarios. If happens, the robot will use the most recent target data and use it for closed loop control until the target is shown again. The eye-in-hand camera feedback is expected to drive the robot-target error less than 2cm, which enables the robot gripper to effectively pick the berries.

\section{Experimental Results and Discussion}
\label{sec: Experimental Results}

In this section, we aim to validate the effectiveness of our proposed algorithms and hardware through a series of laboratory experiments. We will begin with a brief discussion on system identification, which maps pressure space to actuator space. Then, we will examine the resolved-rate algorithm results for position and orientation control, both with and without redundancy resolution, to demonstrate the significance of keeping the actuator variables within their constraints.
Subsequently, we will delve into the topic of simultaneous control of position and orientation. Lastly, we will implement visual servoing using a hybrid of eye-in-hand and eye-to-hand camera measurements, and validate our robot hardware and control algorithm with field tests.

\subsection{System Identification for Soft Actuator}
The process of soft actuator system identification was conducted using a linear guide rail (TS-04-09, igus, USA) and a linear optical encoder (EM1-0-200-N, US Digital, USA). A ramp pressure from 10 psi to 50 psi was applied to the actuator using the xPC target machine, and the resulting change in length was recorded by the optical encoder. The ratio of actuator length to pressure was then calculated for each time-step, and the mean value was used to map actuator length change to pressure ($\frac{l_{i,j}}{P} = 2100^{-1} \left[\frac{m}{psi}\right]$). 

\subsection{Robot Position Control}
\label{subsec: Robot Position Control}
\begin{figure}[t]
    \centering
    \includegraphics[width = 1.00\linewidth]{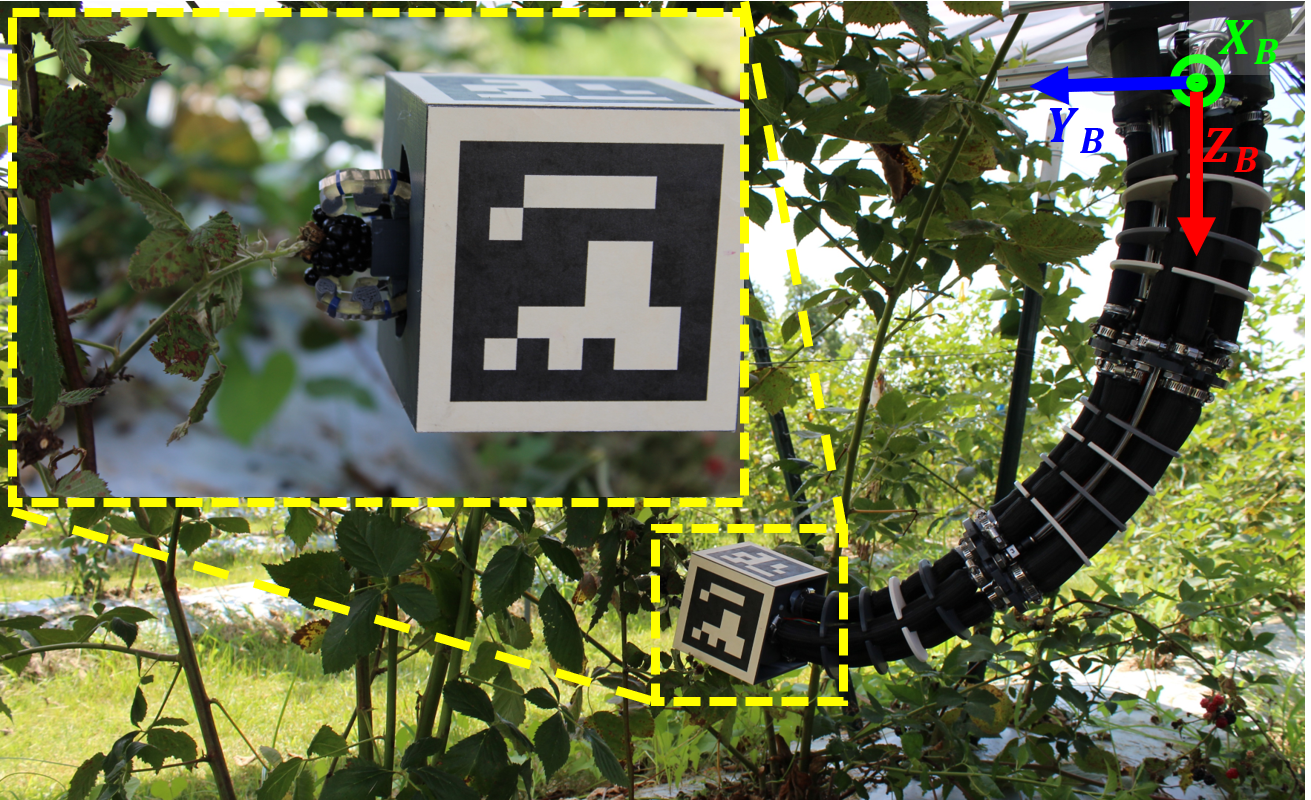}
    \caption{Robotic assembly harvesting a blackberry. The base frame configuration is highlighted where berries are distributed along the Y\_ axis of the robot. The gripper can be seen grasping the berry after successful robot navigation.}
    \label{fig:Harvesting}
    \vspace{-2mm}
\end{figure}

\begin{figure}[tbh]
    \centering
    \includegraphics[width = 1.00\linewidth]{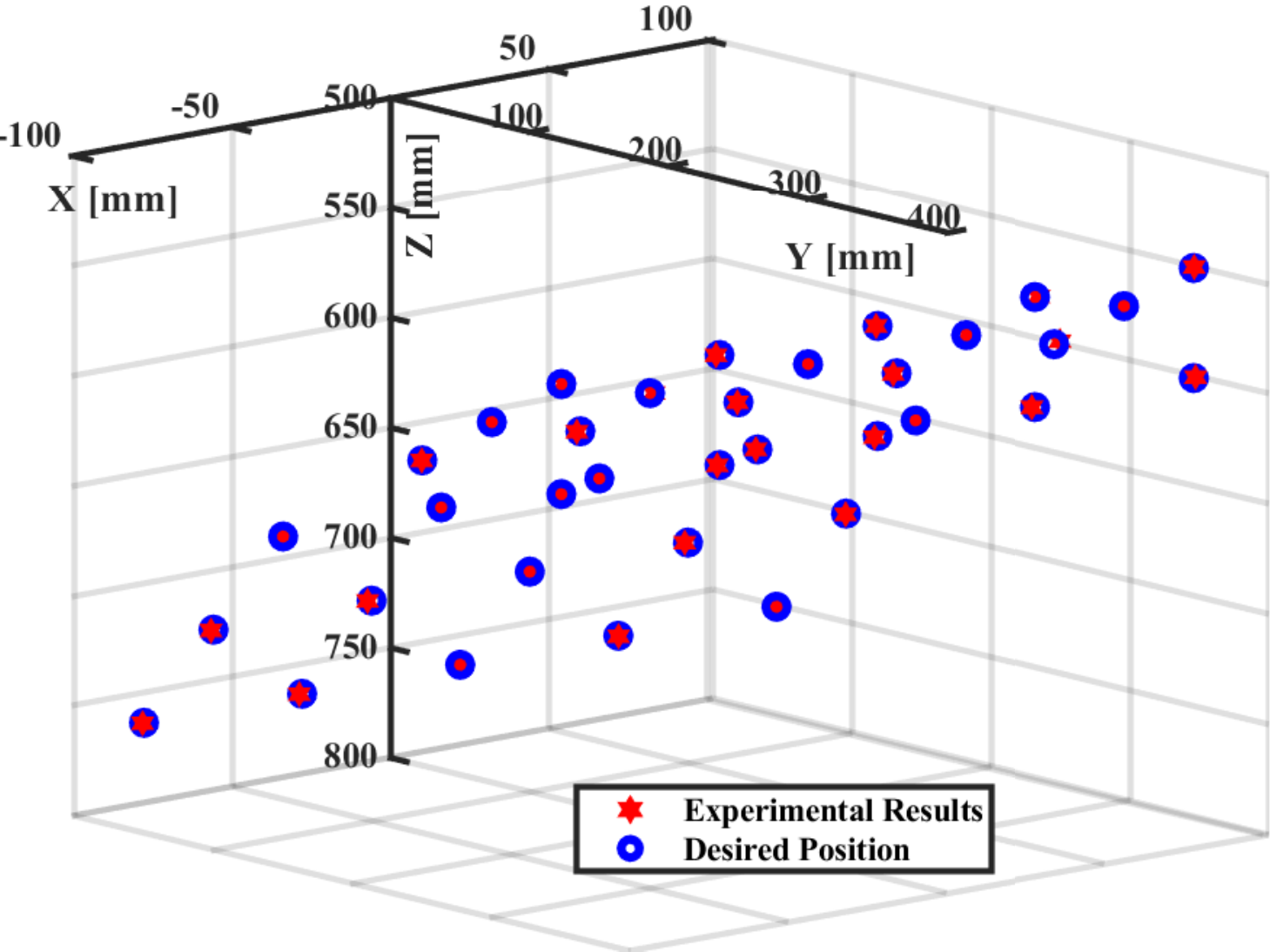}
    \caption{Position control results using redundancy resolution algorithm for joint-limit avoidance.}
    \label{fig:PosData}
    \vspace{-2mm}
\end{figure}

In this study, we defined 35 set-points within the robot's workspace, distributed in a manner that emulates the possible locations of berries in a real harvesting scenario (Fig. \ref{fig:PosData}). The results of our lab study are presented in Fig. \ref{fig:PosData}, with the robot placement shown in Fig. \ref{fig:Harvesting}. Our robot hardware and algorithm achieved a mean error of 0.84\textpm{0.38}mm. It's worth noting that we used a high-resolution optical camera (MicronTracker, Claronav, Toronto, Canada) as the eye-to-hand camera for feedback purposes. This will allow us to eliminate the potential image processing error caused by the ZED-2 camera.

The redundancy resolution algorithm with joint limit avoidance for position control (Task 1 over 3) was defined in the first equation of \eqref{eq: redjoi}, and the following parameters were used
\begin{equation}
\begin{aligned}
        \alpha_2 &= 0.074\\
    \gamma_2 &= \begin{cases}
    0, &  ||e_p||_{2} \leq 30 mm\\
    -0.01, &  ||e_p||_{2} > 30 mm
    \end{cases}
\end{aligned}
\end{equation}
We tuned the gains by trial and error, starting with a small gain for $\alpha_2$ to ensure stable movement in the robot. Then, we tuned $\gamma_2$ to provide enough effect in the resolved-rate algorithm to push the solution to the feasible region, while maintaining a bound where its effects do not change the solution drastically. Finally, we increased $\alpha_2$ to improve the convergence speed, and maintained it in a region that prevents instability. For conventional resolved-rate control, we set $\gamma_2$ to zero at all times, and $\alpha_2 = 0.074$, making the algorithm equivalent to \eqref{eq: res_g}, but only for position control.

\begin{figure}[tbh]
     \centering
     \includegraphics[width=1\linewidth]{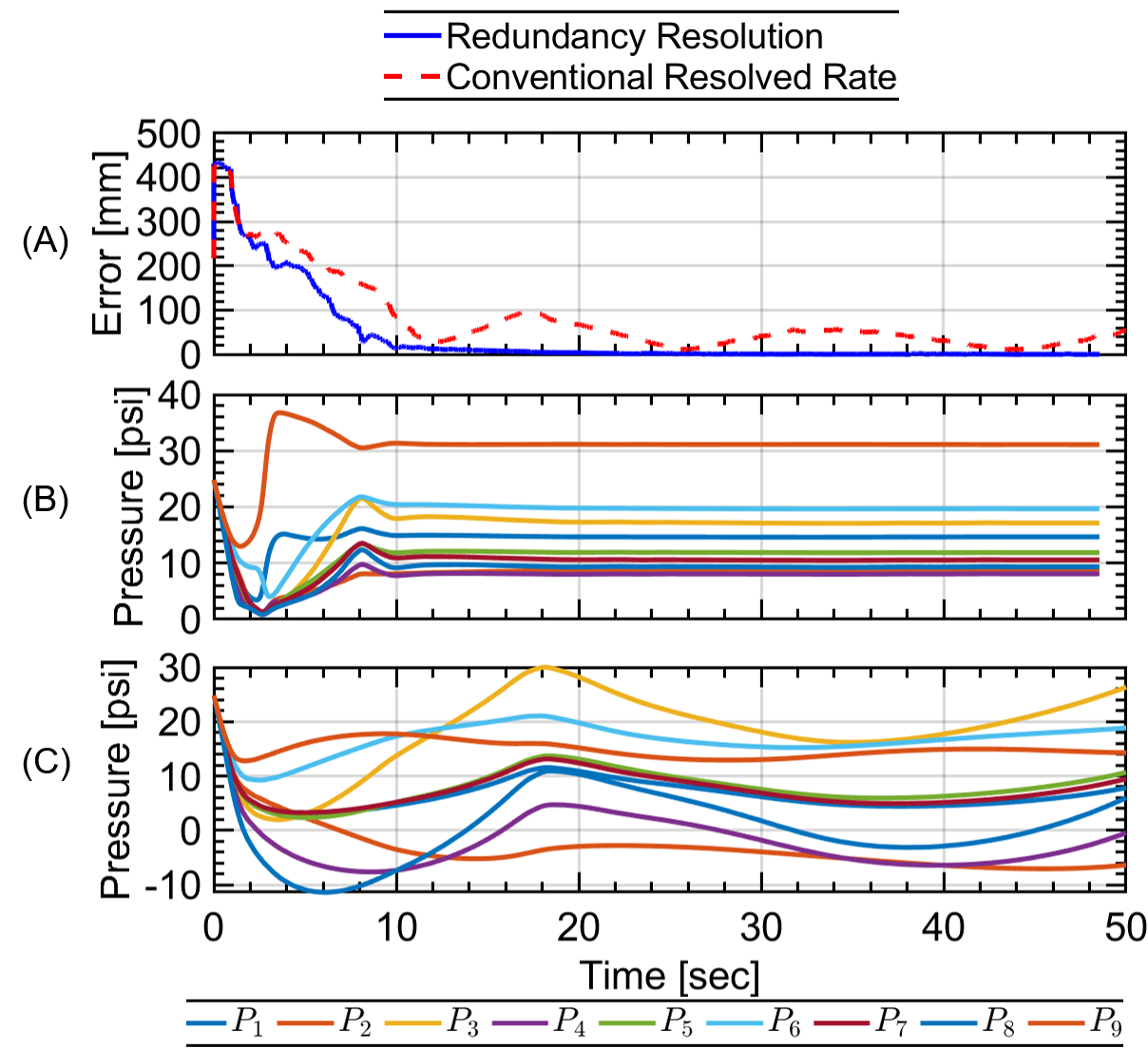}
     \caption{(A) Error comparison for position tracking using redundancy resolution for joint limits avoidance versus conventional resolved-rate algorithm. (B) Soft actuator pressure solution through time generated with the redundancy resolution algorithm with joint limit avoidance. All pressures maintained in physically feasible region. (C) Soft actuator pressure solution through time generated  the conventional resolved-rate control through time. The algorithm has not converged and resulted in infeasible negative pressure solution.
     }
     \label{fig:PosCov}
\end{figure}

To emphasize the significance of joint-limits in achieving both convergence speed and stability, an additional experiment was conducted to compare the response of the proposed redundancy resolution algorithm to the conventional resolved-rate algorithm for position control. As depicted in Fig. \ref{fig:PosCov}, the redundancy resolution algorithm with joint-limits avoidance achieved an error of 4.5 mm after 20 seconds and 0.8 mm error after 30 seconds, while the conventional resolved-rate algorithm was unable to converge, as shown in Fig. \ref{fig:PosCov}A, where the positioning error fluctuated over time. This can be attributed to the infeasible solutions generated by the conventional resolved-rate algorithm without joint limit avoidance. It is important to note that the negative pressures illustrated in Fig. \ref{fig:PosCov}C are infeasible, as the actuators can only elongate, preventing the system from converging. We also want to mention that the experiment was conducted for 100 seconds for both algorithms, and we only present the first 50 seconds data for better readability.

\subsection{Orientation control}
\label{subsec: Robot Orientation Control}
In this part, we validate our proposed approach for orientation control. Specifically, we investigate the performance of the redundancy resolution algorithm with joint-limit avoidance, as described in the second equation of \eqref{eq: redjoi}. To assess the robot's performance, we select 15 data points varying from zero to -150 degrees with respect to the X-axis, while keeping the orientation about the Y- and Z-axes at zero. This is because the robot is symmetric, and thus we only consider rotation about the X-axis.
Our proposed redundancy resolution algorithm with joint-limit avoidance is able to provide an end-effector error of $0.65^{\circ}$ \textpm $0.24^{\circ}$. The gains for our proposed approach for orientation control and joint-limit avoidance (task 2 over 3) using the redundancy resolution algorithm are described as follows
\begin{equation}
    \begin{aligned}
    \alpha_3 &= 0.05\\
    \gamma_3 &= \begin{cases}
    0, & ||e_\zeta||_2 \leq 30^{\circ}\\
    -0.005, & ||e_\zeta||_2 > 30^{\circ}
    \end{cases}
    \end{aligned}
\end{equation}
Similar to the discussion in \ref{subsec: Robot Position Control}, we first set $\alpha_3$ to a small value and tuned $\gamma_3$ to ensure stable and consistent behavior. Then, we increased $\alpha_3$ to improve convergence speed. For the conventional resolved-rate control, we set $\gamma_3$ to zero at all times and $\alpha_3$ to 0.05, which makes it equivalent to \eqref{eq: res_g}, but only for orientation control.

\begin{figure}[t]
     \centering
     \includegraphics[width=1\linewidth]{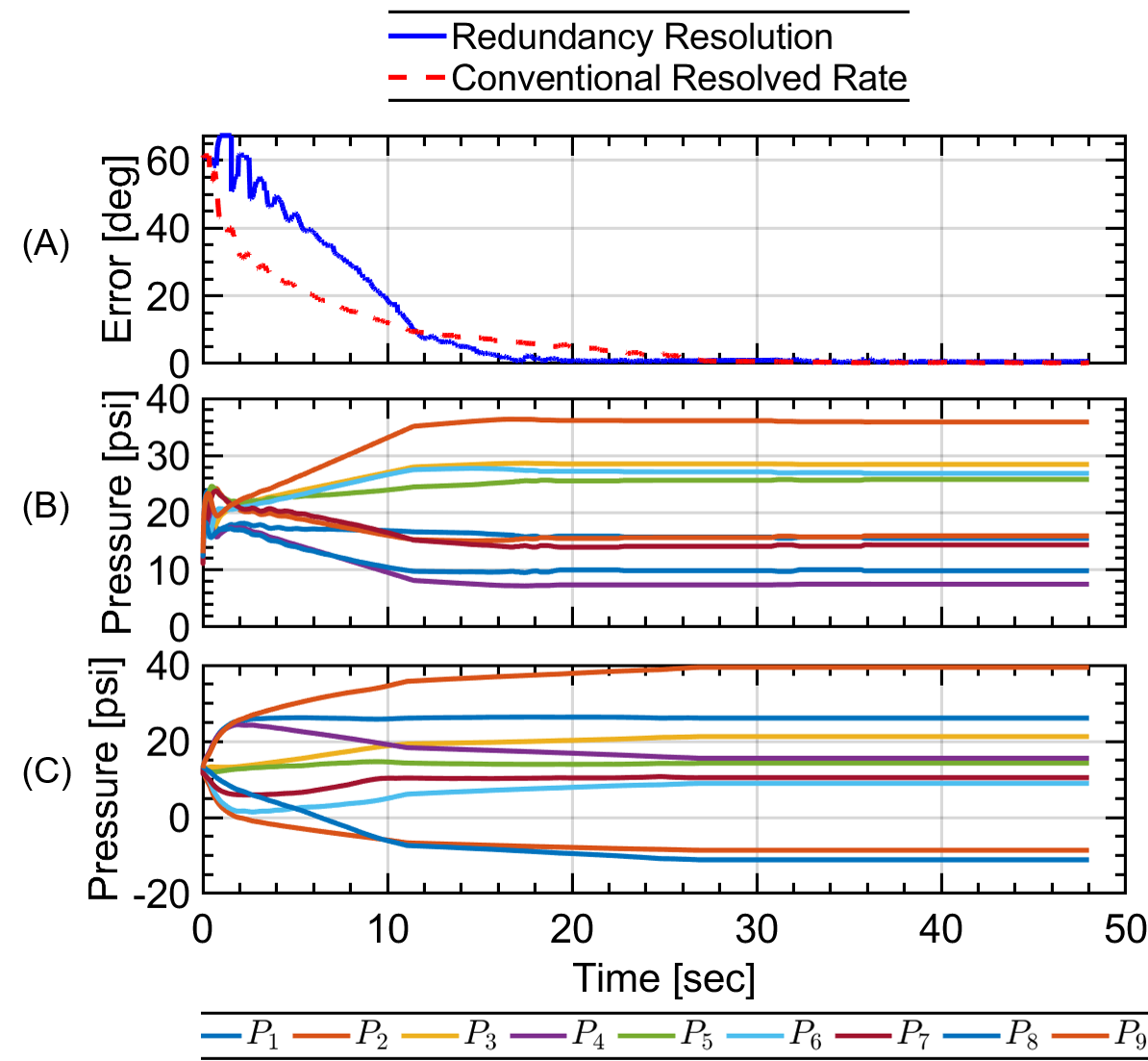}
     \caption{(A) Error comparison for orientation tracking using redundancy resolution for joint limits avoidance versus conventional resolved-rate algorithm. (B) Generated pressure solution through time for redundancy resolution algorithm with joint limit avoidance. All pressures maintained in mechanically feasible region. (C) Generated pressure solution for conventional resolved-rate control through time. The algorithm resulted in slower convergence speed.}
     \label{fig:OriCov}
\end{figure}

To highlight the importance of considering joint-limits, we conducted another experiment to compare the convergence speed and stability of the algorithm with and without joint-limit avoidance. As shown in Fig. \ref{fig:OriCov}, although both algorithms converged, the redundancy resolution algorithm provided a faster response time compared to the conventional resolved-rate algorithm. The redundancy resolution algorithm reached an error of $0.57^{\circ}$ in 17.3 seconds, while the conventional approach reached the same error in 30.9 seconds. 
In Fig. \ref{fig:OriCov} C, it can be observed that the actuator inputs provided negative pressures, resulting in a slower settling speed for the system.

\subsection{ Position and Orientation Control}
\label{subsec: PosOri}
For efficient blackberry harvesting, it is essential to enable simultaneous position and orientation control. To optimize the process, we strategically placed the harvesting hardware to ensure that the berries were scattered along the Y-axis positive direction of the base-frame. As such, controlling the robot orientation along the X and Z axis of the base frame is crucial for successful berry harvesting. Our approach involved prioritizing position control over orientation control, and joint-limit avoidance by incorporating \eqref{eq: rrra}.  In the first part of experiment for the simultaneous position and orientation control, we evaluated the performance of our controller by reaching 7 different poses, achieving an error of 0.64 \textpm{0.27} mm in position and $1.08^{\circ}$ \textpm{1.5}$^{\circ}$ in orientation, specifically in the X and Z axis. The following gains are used for \eqref{eq: rrra} as follows

\begin{equation}
    \begin{aligned}
    \alpha_4 &= 0.05\\
    \gamma_6 &= 1\\
    \gamma_4 &= 0.1\\
    \gamma_5 &= \begin{cases}
    0, & ||e_p||_2 \leq 50 mm\\
    -5e^{-5}, & ||e_p||_2 > 50 mm
    \end{cases}
    \end{aligned}
\end{equation}

To obtain these tuning parameters, $\alpha_4$ was set to a small value to ensure slow but guaranteed convergence. Then, the system was commanded to reach a pose in which all the actuators needed to be extended. $\gamma_6$ was set to 1, and $\gamma_4$ was tuned to be large enough to push the solution into the realizable manifold for position and orientation convergence, without dominating the position control part. After achieving a desirable behavior, the system was commanded to reach another pose in which multiple solutions, including unrealizable solutions, existed. Then, similar to sections \ref{subsec: Robot Position Control} and \ref{subsec: Robot Orientation Control}, position and orientation control were treated as a single task and $\gamma_5$ was tuned. Finally, $\alpha_4$ was increased to improve convergence speed.

\begin{figure}[t]
     \centering
     \includegraphics[width=1\linewidth]{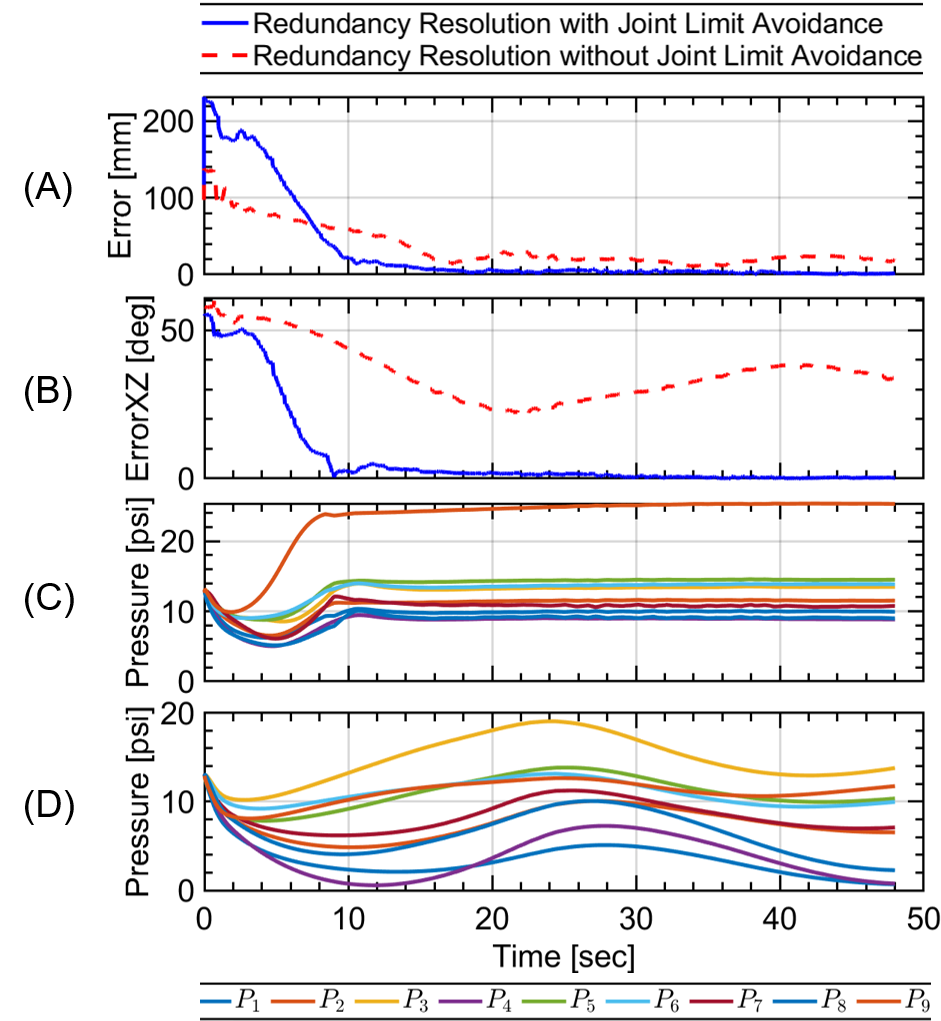}
     \caption{(A) 
     Position error comparison for redundancy resolution algorithm for simultaneous position and orientation control with and without joint-limit avoidance.
     (B) Orientation $L_2$ error comparison around X and Z axis of the base frame. Plot show the results for redundancy resolution algorithm for hybrid position and orientation control with and without joint-limit avoidance.(C) Generated pressure solution for simultaneous position and orientation control with joint-limit avoidance.  (D) Generated pressure solution for simultaneous position and orientation control without joint-limit avoidance.}
     \label{fig:PosOriCov}
\end{figure}

To highlight the superiority of the proposed approach in equation \eqref{eq: rrra} and the importance of maintaining actuator variables within their physically realizable region, another experiment was conducted that compared simultaneous position and orientation control without joint limit avoidance ($\gamma_5$ = 0 at all times). It should be noted that a conventional resolved rate algorithm, as described in equation \eqref{eq: res_g}, was unable to converge to any pose. As shown in Figure \ref{fig:PosOriCov}, both controllers (simultaneous position and orientation control with and without joint-limit avoidance) converged for position due to position prioritization, but the inclusion of joint limits increased the accuracy by a considerable margin. In orientation control, the algorithm with joint-limit avoidance converged to the desired accuracy, whereas the robot was unable to converge to the desired orientation in the other scenario. The reason for this can be seen in Figure \ref{fig:PosOriCov}D, where some actuator variables were below 5 psi, making their elongation negligible.

\subsection{Artificial Berry Harvesting in Benchtop Settings}

\begin{figure}[t]
     \centering
     \includegraphics[width=1\linewidth]{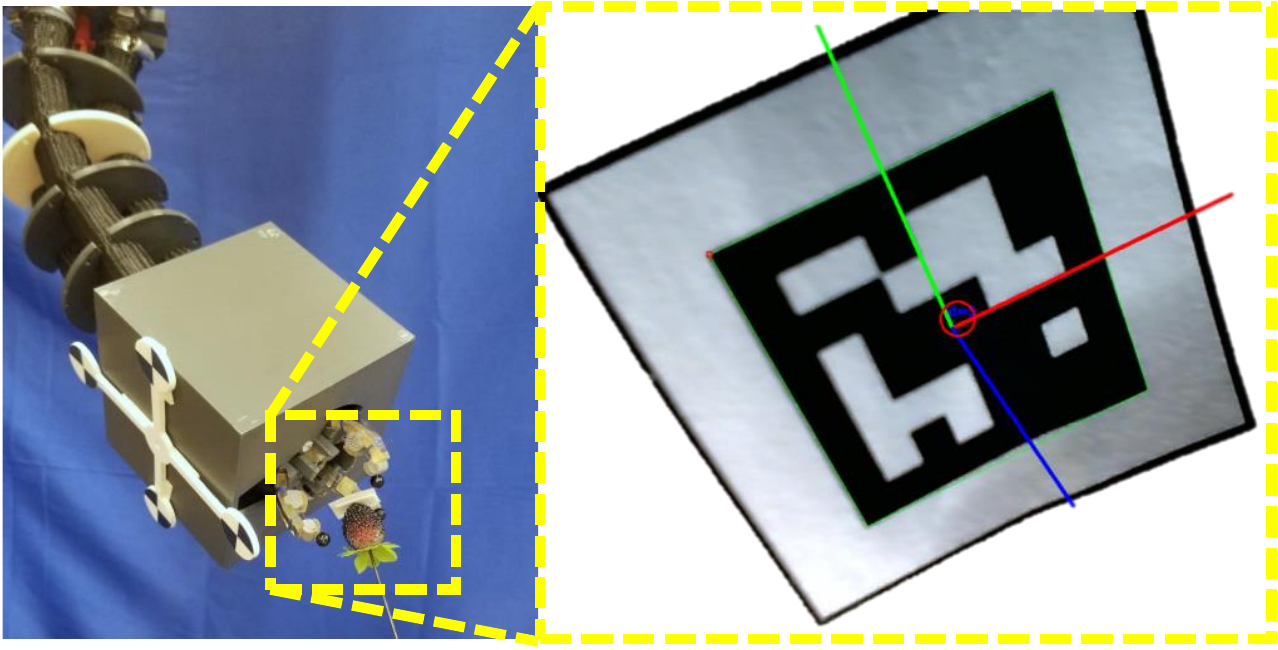}
     \caption{Visual servoing performance for centering the berry, the circle represent the center of the image and the vectors are reference frame of the ArUco tag. The picture is taken when the berry's offset is 3 cm (close to gripper tip) with respect the end-effector frame.}
     \label{fig:Vis1}
\end{figure}

In this section, we evaluate the performance of the proposed berry harvesting workflow using an artificial blackberry. An ArUco tag was attached to the artificial berry to provide the desired position, which is used to drive the robot towards  the target.
The algorithm achieved high accuracy, with a mean error of 0.75 \textpm{0.36} mm for 10 different targets. 
The performance of the fine positioning part in the lab study is depicted in Fig. \ref{fig:Vis1}. As shown in the figure, the ArUco tag was placed in the middle of the eye-in-hand camera imaging frame, which indicates that the gripper was able to place at the proper location for effective harvesting.

\subsection{Field Study}

A field test was conducted at a blackberry research farm in Tifton, GA, USA, to evaluate the proposed platform's performance in a realistic environment. The test involved placing the robot in front of a row of blackberry plants, with the direction of the bushes aligned along the +Y-axis of the base frame, and ripe berries were identified using an eye-to-hand camera with an 80 mm ArUco tag. Fig. \ref{fig:Harvesting} illustrates the execution of a successful harvesting process using the proposed platform. The initial end-effector positioning was performed using a gain-scheduled redundancy resolution resolved-rate algorithm with joint-limit avoidance to ensure accurate positioning of the gripper over the berry. Despite not being able to implement the fine positioning part of the process due to the difficulty of placing small ArUco tags on ripe berries, manual commands were executed to align the gripper with the berry until it was within the robot's grasp workspace. The proposed platform successfully harvested 15 berries and achieved a 100\% success rate. The limited number of tests precludes more extensive statistical analysis, but the results were promising and demonstrate the feasibility of the proposed platform in a challenging and dynamic environment with the bushes' direction aligned along the +Y-axis of the base frame. More detailed results related to the soft gripper harvesting evaluations can be found in \cite{qiu2023gripper}. 
The proposed platform (algorithm and hardware) demonstrated a stable and consistent performance during the harvesting process, highlighting its potential for deployment in agricultural applications.

\section{Conclusion}
\label{sec: Conclusion}
 
This paper proposes a soft continuum system for blackberry harvesting that consists of a pneumatically actuated redundant soft arm and a soft gripper. 
Two general stages were defined for the gripper closed loop control, namely 1) initial placement for the gripper and 2) fine motion control. To achieve accurate position and orientation control for initial placement, a redundancy resolution algorithm was proposed to decouple the position and orientation control with joint-limit avoidance. The detailed parameter tuning method was detailed  to ensure stable, repeatable, and continuous behavior. The proposed algorithm achieved an accuracy of 0.64 ± 0.27 mm and 1.08 ± 1.5° for simultaneous position and orientation control in the lab study. 
Fine positioning was achieved by transforming the eye-in-hand camera measurements to the robot's base frame using the eye-to-hand camera information. 
The approach achieved a 0.75 mm error in the lab study. To ensure stable transitioning between each stage of the harvesting process, a high-level workflow was incorporated, with more focus on the transition between initial placement and fine position stages.
The lab studies successfully translated to a field study, satisfying the berry harvesting requirements. 

Note that this is the preliminary study of the soft robot enabled harvesting. Future work will focus on increasing the process speed by incorporating string rotary encoders introduced in \cite{azizkhani2022dynamic2} and taking into account the dynamic behavior of the robot. Also, locomotion platform will be developed to enable versatile motion within the unstructured terrain environment. Lastly, the robot perception system using the YOLO algorithm will also be incorporated to locate and segment the berries in the eye-to-hand and eye-in-hand cameras.

\section*{Acknowledgment}

The authors would like to thank Benjamin O. Gunderman, Egemen Balban, Brandon Lynchard, and  Huaijin Tu for their contributions to the project.

\ifCLASSOPTIONcaptionsoff
  \newpage
\fi



%
\bibliographystyle{ieeetr}
\bibliography{references}

%










\end{document}